# Learning twin primality from pairs of modular-inverse graphs

Tal Weissblat
Ramat Gan Academic College, Ramat Gan, Israel
Corresponding author: Tal Weissblat
Email: tal.weisblat@gmail.com

## Abstract

This work investigates whether twin-prime pairs can be identified from graph representations derived solely from modular-inverse relations. Each integer is represented as a graph in which vertices correspond to residues modulo that integer and edges encode multiplicative-inverse relations. We provide mathematical justification for two classification approaches: a Siamese Graph Isomorphism Network (SiameseGIN) and logistic regression. We identify the sum of the isolated-vertex proportions of the two graphs as a common structural quantity underlying both approaches and establish conditions under which it separates twin-prime pairs from non-twin pairs. For the SiameseGIN, we analyze the message-passing and pooling operations and show that the resulting graph representations encode the isolated-vertex proportions, allowing its decision score to depend on the same structural quantity. For logistic regression, we show that the decision score can be expressed as an affine function of this quantity and therefore admits perfect separation under these conditions. We further establish conditions under which the validation and test ranges admit a common separating threshold, providing a theoretical explanation for the empirical observation that the validation-selected threshold also achieves perfect separation on the test set. We evaluate both approaches experimentally using disjoint training, validation, and test ranges. A threshold selected exclusively on the validation set is applied unchanged to the test set, where both logistic regression and the SiameseGIN achieve a balanced accuracy of 1.000, with no false-positive or false-negative classifications. Additional experiments demonstrate perfect classification across extended numerical ranges, including integers near $10^6$, without retraining or threshold adjustment. The principal contribution is a mathematical characterization of how the SiameseGIN architecture processes modular-inverse graphs for integers below $10^4$, while the experimental results demonstrate agreement with the theoretical analysis.

**Keywords:** twin primes; modular-inverse graphs; graph neural networks; SiameseGIN; logistic regression; graph classification; number theory; primality classification.

2020 Mathematics Subject Classification. Primary: 68T07; Secondary: 05C90, 11A41.

# 1 Introduction

Prime numbers are fundamental objects in number theory, and understanding their distribution remains one of the central problems in mathematics [1–3]. Of particular interest are twin primes, pairs of prime numbers that differ by two, such as (5, 7) [2]. The twin prime conjecture, discussed in the classical work of Hardy and Littlewood [4], states that infinitely many such pairs exist. Major advances in understanding prime gaps include the work of Goldston, Pintz, and Yıldırım on small gaps [5], followed by breakthroughs establishing and improving bounded gaps between primes [6-8]. Recent developments concerning large gaps between primes are surveyed by Rassias [9]. Nevertheless, the twin prime conjecture itself remains unresolved.

Recent developments in machine learning have introduced new computational approaches for studying mathematical structures and identifying patterns in mathematical data [10]. In particular, graph neural networks (GNNs) provide a natural framework for learning from objects represented as graphs by combining graph structure with learned representations [11]. The Graph Isomorphism Network (GIN) is an expressive message-passing GNN architecture designed to distinguish graph structures through neighborhood aggregation [12, 13]. Previous research has investigated the application of machine learning to prime-number distributions [14]. In contrast, the present work studies twin-prime classification through graph representations of integers.

In our previous work, we introduced modular-inverse graphs as graph representations of integers and investigated their use for learning primality [15]. The present study extends this framework from the classification of individual integers to the classification of integer pairs, with a particular focus on twin primes.

In this work, we study twin-prime classification through modular-inverse graphs. Each integer $n$ is represented by a graph $G_n$, whose structure is determined by multiplicative inverses modulo $n$. A candidate pair $(n, n+2)$ is consequently represented by the graph pair $(G_n, G_{n+2})$, allowing twin primality to be formulated as a graph-pair classification problem. A SiameseGIN with shared parameters is used to process the two graphs and classify the resulting pair. This architecture is motivated by the fact that twin primality requires two separate primality decisions, while shared parameters ensure consistent processing of both graphs. GIN was selected because its neighborhood aggregation captures the distinction between isolated and non-isolated vertices, a key structural property of modular-inverse graphs underlying the proposed classification method.

The study combines this graph-learning approach with a mathematical analysis of the underlying graph structure. In particular, we examine the normalized number of isolated vertices, $\rho(n)$, and establish conditions under which twin-prime and non-twin pairs are linearly separable. We further establish conditions under which the validation and test ranges admit a common separating threshold.

The proposed approach is evaluated using consecutive and non-overlapping training, validation, and test ranges. In addition to the SiameseGIN, logistic regression based on the theoretically motivated graph features is considered. This allows the empirical behavior of the learned graph model to be compared with a simpler classifier directly connected to the mathematical analysis. The results demonstrate agreement between the theoretical structural characterization and the observed classification behavior.

The main contributions of this work are threefold. First, we formulate twin-prime classification as a graph-pair learning problem using pairs of modular-inverse graphs and provide a mathematical characterization of how the SiameseGIN architecture processes these graphs, showing that its graph-level representations depend on the proportions of isolated vertices. Second, we derive the corresponding decision score and establish sufficient conditions for linear separation and for the existence of a common separating threshold across bounded numerical ranges. Third, we experimentally evaluate the SiameseGIN and a theoretically motivated logistic-regression classifier, demonstrating agreement between the theoretical analysis and the observed classification results.

## 2 Theoretical Analysis

This section develops the mathematical framework used for twin-prime classification with pairs of modular-inverse graphs. We begin with the main number-theoretic and graph-theoretic definitions, followed by the structural properties of modular-inverse graphs introduced in our previous work [15]. For completeness, we state and provide a full proof of the structural characterization used in the present work. These results are then extended from individual integers to pairs of the form $(n, n+2)$, providing the basis for the theoretical analysis of twin-prime classification.

### 2.1 Definitions

The standard number-theoretic definitions and results used in this section follow Burton [16, Secs. 2.3, 4.2, 7.2]. We first introduce the main mathematical objects used throughout this work. For an integer $n \geq 2$, let $Z_n = \{0, 1, \dots, n-1\}$ denote the set of integers modulo $n$. An element $a \in Z_n$ has a multiplicative inverse modulo $n$ if there exists $b \in Z_n$ such that $ab \equiv 1(mod\ n)$. Such an inverse exists if and only if $gcd(a, n) = 1$. Therefore,

$$Z_n^* = \{a \in Z_n \,:\, gcd(a, n) = 1\}$$

The number of elements in $Z_n^*$ is given by Euler's totient function

$$\varphi(n) = |Z_n^*|$$

In particular, if $p$ is prime, then $\varphi(p) = p - 1$. Among these elements, some are their own multiplicative inverses. We denote their number by $s(n)$, so that

$$s(n) = |\{a \in Z_n^* \,:\, a^2 \equiv 1\ (mod\ n)\}|$$

Using these definitions, we associate each integer $n \geq 2$ with a modular-inverse graph

$$G_n = (V_n, E_n)$$

Its vertex set is $V_n = Z_n$, and two vertices $a$ and $b$ are connected whenever $ab \equiv 1(mod\ n)$. The graph is undirected, and self-loops are included when $a = b$. A vertex is isolated if it has no incident edge, including no self-loop. To describe the structure of $G_n$, let $I(n)$ denote the number of isolated vertices in the graph. We then define the normalized isolated-vertex ratio as

$$\rho(n) = \frac{I(n)}{n}$$

These objects provide the basis for the structural analysis of prime and composite integers and, subsequently, for the study of twin-prime pairs.

### 2.2 Structural Characterization

The following theorem and its proof are reproduced from our previous work [15] to make the present analysis self-contained. This result provides the theoretical foundation for the analysis developed in this work.

**Theorem 1.** (Prime–Composite Graph Distinction [15]).
Let $G_n$ be the modular-inverse graph associated with an integer $n \geq 2$. Then:

1. Every edge connecting two distinct vertices is isolated from every other edge. Hence, all such edges are pairwise disjoint.
2. If $n > 2$ is prime, then the corresponding graph has exactly $(n-3)/2$ edges connecting distinct vertices and no isolated nonzero vertices.
3. If $n > 2$ is composite, then the corresponding graph has exactly $(\varphi(n) - s(n))/2$ edges connecting distinct vertices and $n - 1 - \varphi(n)$ isolated nonzero vertices.

**Proof.** For (1), let $\{a, b\}$ be an edge connecting two distinct vertices of $G_n$. Then $ab \equiv 1\ (mod\ n)$. Suppose that $a$ is also connected to a vertex $c$. Then $ac \equiv 1\ (mod\ n)$. Multiplying by $b$ gives $abc \equiv b\ (mod\ n)$. Since $ab \equiv 1\ (mod\ n)$, it follows that $c \equiv b\ (mod\ n)$, and therefore $c = b$ in $Z_n$. Thus $a$ cannot be connected to any other vertex. The same holds for b. Therefore, (1) follows.

For (2), suppose that $n > 2$ is prime. By Euler's totient formula, $\varphi(n) = n - 1$, so every nonzero vertex is invertible. If $a$ is self-inverse, then $a^2 \equiv 1\ (mod\ n)$, and hence $(a-1)(a+1) \equiv 0\ (mod\ n)$. Since $n$ is prime, it follows that $a \equiv 1\ (mod\ n)$ or $a \equiv -1\ (mod\ n)$. Thus, the only self-inverse nonzero vertices are $1$ and $n - 1$. The remaining $n - 3$ vertices therefore form disjoint pairs of multiplicative inverses, giving exactly $(n-3)/2$ edges connecting distinct vertices. Since every nonzero vertex is invertible, there are no isolated nonzero vertices. Therefore, (2) follows.

For (3), suppose that $n > 2$ is composite. By definition, $\varphi(n)$ is the number of invertible vertices, while $s(n)$ is the number of self-inverse vertices. Thus, $\varphi(n) - s(n)$ invertible vertices have distinct multiplicative inverses. By (1), these vertices form disjoint pairs, giving exactly $(\varphi(n) - s(n))/2$ edges connecting distinct vertices. Among the $n - 1$ nonzero vertices, $\varphi(n)$ are invertible, so the remaining $n - 1 - \varphi(n)$ vertices are isolated. Therefore, (3) follows.

### 2.3 Twin-Prime Classification Problem

We now extend the graph-based formulation from single integers to pairs of integers. For each $n \geq 2$, we consider the pair $(n, n+2)$ and the corresponding pair of modular-inverse graphs $(G_n, G_{n+2})$. Our goal is to determine whether both $n$ and $n + 2$ are prime. This gives the following classification problem.

**Problem 1** (Twin-Prime Classification).
Given an integer $n \geq 2$, consider the pair $(n, n+2)$ and the corresponding modular-inverse graphs $G_n$ and $G_{n+2}$. Determine whether both graphs correspond to prime integers, equivalently, whether $(n, n+2)$ forms a twin-prime pair.

By Theorem 1, $n$ and $n + 2$ are both prime if and only if neither $G_n$ nor $G_{n+2}$ contains an isolated nonzero vertex. Thus, twin-prime classification can be expressed entirely in terms of the structural properties of the corresponding pair of modular-inverse graphs.

Since $I(n) = n - \varphi(n)$ and $\rho(n) = I(n)/n$, the condition for primality is equivalent to:

$$n\rho(n) = 1$$

Consequently, the pair $(n, n + 2)$ forms a twin-prime pair if and only if both conditions hold:

$$\begin{aligned} n\rho(n) &= 1 \\ (n+2)\rho(n+2) &= 1 \end{aligned}$$

These conditions define an exact arithmetic baseline that requires neither model training nor threshold optimization.

### 2.4 Pair-Level Structural Characterization

We next examine how the structural properties of individual modular-inverse graphs extend to pairs $(G_n, G_{n+2})$. For n ≥ 2, the set of vertices of $G_n$ is

$$V_n = \{0, 1, \ldots, n - 1\}$$

Each vertex $a \in V_n$ belongs to exactly one of three structural categories:

I. Distinct-inverse vertex. In this case, $a$ is connected to the distinct vertex $a^{-1}$, equivalently, $gcd(a, n) = 1$ and $a^2 \not\equiv 1(mod\, n)$.

II. Self-inverse vertex. In this case, $a = a^{-1}$ and therefore has a self-loop, equivalently $a^2 \equiv 1(mod\, n)$.

III. Isolated vertex. In this case, there exists no $b \in V_n$ such that ab ≡ 1 (mod n), equivalently $gcd(a, n) > 1$.

The purpose of the remainder of this section is to introduce the isolated-vertex ratio $\rho(n)$, which will serve as a key structural quantity in the subsequent analysis. For an integer $n \geq 2$, let $I(n)$ denote the number of isolated vertices in $G_n$. From Theorem 1, the only isolated vertex when $n$ is prime is $0$, whereas for composite $n$, additional nonzero vertices are isolated. Hence, $I(n) = n - \varphi(n)$. It is useful to normalize this quantity by the size of the graph. Define the isolated-vertex ratio

$$\rho(n) = \frac{I(n)}{n}$$

Since $I(n) = 1$ when $n$ is prime and $I(n) > 1$ when $n$ is composite, we conclude $\rho(n) = 1/n$ for prime number $n$ while $\rho(n) \geq 2/n$ for a composite number. We now extend this quantity to the pair of modular-inverse graphs associated with $n$ and $n + 2$. Thus, the pair-level representation yields the following structural characterization:

- Twin prime $\rho(n) = \frac{1}{n}$ and $\rho(n + 2) = \frac{1}{n+2}$
- Negative pairs $\rho(n) \geq \frac{2}{n}$ or $\rho(n + 2) \geq \frac{2}{n+2}$

## 2.5 Mathematical Separation of Positive and Negative Instances

In this section we introduce a model-independent quantity that provides the common mathematical basis for both the SiameseGIN and logistic regression. Based on the isolated-vertex ratio $\rho(n)$, define

$$D(n) = \rho(n) + \rho(n+2)$$

This quantity plays a dual role in our analysis. First, $D(n)$ encodes whether the pair $(n, n+2)$ is a twin-prime pair or not. Second, it is directly related to the decision scores of both classification models. Under the conditions considered below, these scores take the forms

$$\begin{aligned} S_{\mathrm{GIN}}(n) &= \alpha D(n) + \gamma \\ S_{\mathrm{LR}}(n) &= -D(n) + b \end{aligned}$$

For the SiameseGIN, the first expression follows under the equal-coefficient condition, while for logistic regression the second corresponds to an equal-weight linear score. In both cases, $D(n)$ therefore determines the decision score up to the model parameters. The sign of the score determines the predicted class, and setting the score equal to zero defines the corresponding decision hyperplane. We further establish conditions under which the validation and test ranges admit a common separating threshold for $D(n)$.

Throughout in what follows, an integer $n \geq 2$ is referred to as positive if both $n$ and $n+2$ are prime, and as negative otherwise. Thus, a positive $n$ corresponds to a twin-prime pair $(n, n+2)$.

**Lemma 1** (Composite Lower Bound).
For every composite integer $n$,

$$\rho(n) \geq \frac{1}{\sqrt{n}} - \frac{1}{n}$$

**Lemma 2** (Separation of Twin-Prime Pairs).
Let $[N_1, N_2]$ be such that $N_1 > 16$ and $N_2 < N_1^2/16 - 2$. Then one has

$$\max_{\substack{n \in [N_1, N_2] \\ Positive}} D(n) < \min_{\substack{n \in [N_1, N_2] \\ Negative}} D(n)$$

**Lemma 3** (Existence of a Common Validation–Test Threshold).
Let the validation pairs be defined over $n \in [N_1, N_2]$ and the test pairs over $n \in [N_3, N_4]$, with $N_2 < N_3$, where both intervals satisfy the conditions of Lemma 2. If

$$N_4 < N_1^2/16 - 2$$

then there exists a threshold $T_D$ such that, for both the validation and test ranges, $D(n) < T_D$ for every positive instance, and $D(n) > T_D$ for every negative instance. Consequently, the same threshold $T_D$ perfectly separates positive and negative instances on both ranges.

**Proof of Lemma 1.** Let $p$ be the smallest prime divisor of the composite integer $n$. Since $n$ is composite, $p \leq \sqrt{n}$. Every proper positive multiple of $p$,

$$p, 2p, \ldots, (n/p - 1)p,$$

has a nontrivial common divisor with $n$ and therefore corresponds to an isolated vertex of $G_{\mathrm{n}}$. Hence, $I(n) \geq n/p - 1$. Using $\rho(n) = I(n)/n$, we obtain

$$\begin{aligned} \rho(n) &\geq (n/p - 1)/n \\ &= 1/p - 1/n \end{aligned}$$

Since $p \leq \sqrt{n}$, we have $1/p \geq 1/\sqrt{n}$. Therefore, $\rho(n) \geq 1/\sqrt{n} - 1/n$.

**Proof of Lemma 2.** If $n$ is positive, then both $n$ and $n + 2$ are prime. Hence

$$\begin{aligned} \mathrm{D(n)} &= \frac{1}{\mathrm{n}} + \frac{1}{\mathrm{n} + 2} \\ &\leq \frac{1}{\mathrm{N}_1} + \frac{1}{\mathrm{N}_1 + 2} \\ &< \frac{2}{N_1} \end{aligned}$$

Now let $n$ be negative. Then at least one of $n$ and $n + 2$ is composite. Let this composite integer be $m \in \{n, n + 2\}$. Since $m \leq N_2 + 2$, and $\rho(m) \geq 1/\sqrt{\mathrm{m}} - 1/m$ by Lemma 1 for every composite $m$, we obtain

$$\begin{aligned} D(n) &\geq \rho(m) \\ &\geq 1/\sqrt{\mathrm{m}} - 1/m \\ &\geq 1/\sqrt{\mathrm{N}_2 + 2} - 1/(N_2 + 2) \end{aligned}$$

where the last inequality follows because the function $f(x) = 1/\sqrt{x} - 1/x$ is decreasing for $x > 4$ and $m \leq N_2 + 2$. The assumption

$$N_2 < \frac{N_1^2}{16} - 2$$

implies

$$N_2 + 2 < \frac{N_1^2}{16}.$$

Since $f(x) = 1/\sqrt{x} - 1/x$ is decreasing for $x > 4$, we obtain

$$\begin{aligned} \frac{1}{\sqrt{N_2 + 2}} - \frac{1}{N_2 + 2} &> \frac{4}{N_1} - \frac{16}{N_1^2} \\ &> \frac{3}{N_1} \\ &> \frac{1}{N_1} + \frac{1}{N_1 + 2} \end{aligned}$$

where the second inequality follows from $N_1 > 16$. It follows that for every positive $n$, $D(n) \leq 1/N_1 + 1/(N_1+2)$, whereas, for every negative $n$,

$D(n) \geq \frac{1}{\sqrt{N2+2}} - \frac{1}{N_2+2}$. Consequently,

$$\max_{\substack{n\in[N_1,N_2]\\ Positive}} D(n) < \min_{\substack{n\in[N_1,N_2]\\ Negative}} D(n)$$

**Proof of lemma 3.** For the validation interval, Lemma 2 gives the guaranteed threshold interval

$$I_{val} = \left(\frac{1}{N_1} + \frac{1}{N_1 + 2}, \frac{1}{\sqrt{N_2 + 2}} - \frac{1}{N_2 + 2}\right) =: (L_{val}, U_{val})$$

$$I_{test} = \left(\frac{1}{N_3} + \frac{1}{N_3 + 2}, \frac{1}{\sqrt{N_4 + 2}} - \frac{1}{N_4 + 2}\right) =: (L_{test}, U_{test})$$

Since $N_3 > N_1$, it follows that $L_{test} < L_{val}$. Therefore, to guarantee that $I_{val}$ and $I_{test}$ overlap, it is sufficient to show that $L_{val} < U_{test}$. By the assumptions $N_1 > 16$ and $N_4 < N_1^2/16 - 2$, the same argument used in Lemma 2 gives

$$1/N_1 + \frac{1}{N_1 + 2} < 1/\sqrt{(N_4 + 2)} - \frac{1}{N_4 + 2}$$

Hence, $L_{val} < U_{test}$, and therefore $I_{val} \cap I_{test} \neq \emptyset$. Consequently, there exists a threshold that perfectly separates positive and negative instances on both the validation and test sets. This proves the result.

Lemma 3 guarantees the existence of a common separating threshold for the validation and test ranges. However, it does not guarantee that every threshold maximizing balanced accuracy on the validation set will also perfectly separate the test set. In our experiments, the classification threshold is selected exclusively on the validation set and applied unchanged to the test set. Its performance on the test set is therefore an empirical result rather than a direct consequence of Lemma 3.

## 2.6 SiameseGIN - Mathematical Justification

The purpose of this subsection is to derive the decision score of the SiameseGIN and establish its connection to the separating quantity $D(n)$ introduced in Section 2.5. We first show how the isolated-vertex ratios $\rho(n)$ and $\rho(n+2)$ determine the graph-level representations and, consequently, the final logit-difference score. We then show that, under the equal-coefficient condition, this score takes the form

$$S_{\mathrm{GIN}}(n) = \alpha D(n) + \gamma$$

Combined with the separation property of $D(n)$ established in Lemma 2, this provides the mathematical justification for the ability of the SiameseGIN to distinguish positive and negative instances.

The SiameseGIN model can be viewed as consisting of two main phases: embedding and classification. In the embedding phase, the two input graphs $G_n$ and $G_{n+2}$ are independently processed by the same GIN model with shared parameters. For each graph, GIN produces vertex embeddings through its aggregation rule, which are then combined by global mean pooling to obtain the graph-level representations $z(G_n)$ and $z(G_{n+2})$. In the classification phase, these representations are concatenated into the pair representation

$$z_n = [z(G_n) \,||\, z(G_{n+2})]$$

and passed through a linear classifier, which produces two logits $s_0(z_n)$ and $s_1(z_n)$, corresponding to the negative and positive classes, respectively. The classifier score is then defined as the difference $S_{\text{GIN}}(z_n) = s_1(z_n) - s_0(z_n)$. Since the classifier is linear, the score is an affine function of $z_n$, and can therefore be written in the general form

$$\begin{aligned} S_{\text{GIN}}(z_n) &= s_1(z_n) - s_0(z_n) \\ &= w^T z_n + b \end{aligned}$$

where $w$ and $b$ are determined by the parameters of the linear classifier. Consequently,

$$H = \{z \in R^{2h} : w^T \text{z} + b = 0\}$$

where $h$ is the graph-embedding dimension and $w \neq 0$. The two sides of this hyperplane correspond to the two predicted classes: $S_{\text{GIN}}(z_n) > 0$ gives a positive prediction, while $S_{\text{GIN}}(z_n) < 0$ gives a negative prediction. For notational convenience, since $z_n$ is uniquely determined by the input pair $(G_n, G_{n+2})$, we will also write $S_{\text{GIN}}(n) = S_{\text{GIN}}(z_n)$.

**Theorem 2.** The SiameseGIN decision score has the form

$$S_{\text{GIN}}(n) = \alpha\rho(n) + \beta\rho(n+2) + \gamma$$

where $\alpha, \beta, \gamma$ are constants determined by the trained model parameters. For $\alpha = \beta$, this reduces to

$$S_{\text{GIN}}(n) = \alpha D(n) + \gamma$$

Furthermore, if $\alpha = \beta < 0$ and $n \in [N_1, N_2]$, where $N_1 > 16$ and $N_2 < N_1^2/16 - 2$, then the score ranks every positive instance strictly above every negative instance, independently of $\gamma$, and consequently AUC = 1.

**Experimental setting.** In the implemented SiameseGIN, the two graph embeddings are processed by shared GIN layers and combined using a final linear classifier with tied weights. Specifically, the same classifier weight matrix is applied separately to each graph embedding, and the resulting outputs are added with a single bias term. This architectural constraint enforces α = β by construction, independently of training. The condition α < 0, however, is not enforced and was verified after training. For the selected trained model, $\alpha = \beta = -24.2369 < 0$, satisfying the coefficient assumptions of Theorem 2.

Moreover, the validation and test ranges satisfy the conditions of Lemma 3, which guarantees the existence of a common threshold $T_D$ for $D(n)$. Since $S_{\mathrm{GIN}}(n) = \alpha D(n) + \gamma$ with $\alpha < 0$, this induces a corresponding common threshold for the SiameseGIN decision score. This is consistent with the empirical observation that the validation-selected threshold also achieved perfect separation on the test set.

To establish Theorem 2, we first derive the following lemma characterizing the GIN representation.

**Lemma 4** (GIN Representation of $G_n$).
Let a GIN be applied to the modular-inverse graph $G_n$, where all vertices have the same initial feature. After the GIN aggregation layers, all isolated vertices have the same final embedding $z_I$, while all non-isolated vertices have the same final embedding $z_N$. Consequently, global mean pooling yields

$$\mathrm{z}(\mathrm{G_n}) = \rho(n) z_I + \big(1 - \rho(n)\big) z_N$$

where $\rho(n)$ denotes the fraction of isolated vertices in $G_n$.

**Proof of Lemma 4.** Let $G_n = (V_n, E_n)$ be a modular-inverse graph. Since no external node information is provided to the model, every vertex $v \in V_n$ is initialized with the same feature, $h_v^0 = [1]$. The node representations are then updated using the GIN aggregation rule [13]. At layer $k$, the representation of a vertex $v$ is given by

$$h_v^k = \mathrm{ReLU}\left(\mathrm{MLP_k}\left((1 + \varepsilon_k) h_v^{k-1} + \textstyle\sum_{u \in N(v)} h_u^{k-1}\right)\right)$$

where $N(v)$ denotes the set of neighbors of $v$, including $v$ itself when a self-loop is present (counted as one message), $\varepsilon_k$ controls the contribution of the vertex's own representation and is fixed at zero in both layers, and $\mathrm{MLP_k}$ is the multilayer perceptron associated with the $k$-th GIN layer.

To understand how this update acts on $G_n$, we first consider the aggregation term separately. Since $h_v^0 = [1]$ for every vertex, the first-layer aggregation is determined entirely by $|N(v)|$, the number of neighbors of $v$. Specifically,

$$\begin{aligned} a_v^1 &= (1 + \varepsilon_1) h_v^0 + \sum\nolimits_{u \in N(v)} h_u^0 \\ &= (1 + \varepsilon_1)[1] + |N(v)|[1] \\ &= (1 + \varepsilon_1 + |N(v)|)[1] \end{aligned}$$

Although the graph contains three structural vertex types, vertices connected to distinct inverse vertices and self-inverse vertices with self-loops each contribute exactly one neighbor message. Since all initial node features are identical, these vertices have identical embeddings at every GIN layer. Thus, only two embedding cases need to be considered.

We now distinguish between two cases according to whether v is isolated or non-isolated in $G_n$.

**Case 1** ($v$ is isolated).

First GIN layer. If $v$ is an isolated vertex of $G_n$, then it has no neighbors and therefore $|N(v)| = 0$. Hence, the first-layer aggregation becomes $a_v^1 = [1]$. Applying the first MLP followed by ReLU gives

$$h_v^1 = \mathrm{ReLU}(\mathrm{MLP}_1([1]))$$

Therefore, since all isolated vertices have the same initial feature and no neighboring contributions, they all receive the same representation after the first GIN layer.

Second GIN layer. Since $v$ is isolated, it again receives no neighboring contributions in the second GIN layer. Therefore, $a_v^2 = h_v^1$. Applying the second MLP followed by ReLU gives

$$h_v^2 = \mathrm{ReLU}(\mathrm{MLP}_2(h_v^1))$$

Substituting $h_v^1$ gives the embedding vector

$$h_v^2 = \mathrm{ReLU}(\mathrm{MLP}_2(\mathrm{ReLU}(\mathrm{MLP}_1([1])))) \tag{1}$$

Thus, $h_v^2$ is the final representation of an isolated vertex before mean pooling.

**Case 2** ($v$ is non-isolated).
First GIN layer. If $v$ is non-isolated, then it participates in a modular-inverse relation. Thus, the aggregation contains a contribution from its inverse neighbor. Since every vertex has the initial feature [1], the first-layer aggregation is

$$\begin{aligned} a_v^1 &= [1] + [1] \\ &= 2[1] \end{aligned}$$

Applying the first MLP followed by ReLU gives

$$h_v^1 = \mathrm{ReLU}(\mathrm{MLP}_1(2[1]))$$

Thus, all non-isolated vertices receive the same representation after the first GIN layer.

Second GIN layer. Since $v$ is non-isolated, it again receives a contribution from its inverse neighbor $u$. After the first GIN layer, both $v$ and $u$ have the same representation $h_v^1$. Therefore,

$$\begin{aligned} a_v^2 &= h_v^1 + h_v^1 \\ &= 2h_v^1 \end{aligned}$$

Applying the second MLP followed by ReLU gives

$$h_v^2 = \mathrm{ReLU}(\mathrm{MLP}_2(2h_v^1)).$$

Substituting $h_v^1$ gives the embedding vector

$$h_v^2 = \mathrm{ReLU}(\mathrm{MLP}_2(2\,\mathrm{ReLU}(\mathrm{MLP}_1(2[1])))). \tag{2}$$

Thus, all non-isolated vertices receive the same final representation before mean pooling.

**Mean pooling.** Let $I(n)$ denote the number of isolated vertices in $G_n$. Since the latter contains $n$ vertices, there are $I(n)$ isolated vertices and $n - I(n)$ non-isolated vertices. Therefore, mean pooling gives

$$\begin{aligned} z(G_n) &= \frac{\left[I(n)z_I + \left(n - I(n)\right)z_N\right]}{n} \\ &= \rho(n)z_I + \left(1 - \rho(n)\right)z_N \end{aligned}$$

where $z_I$ and $z_N$ denote the final representations of isolated and non-isolated vertices, respectively, as given by Eqs. (1) and (2). Since $\rho(n) = I(n)/n$, we obtain $z(G_n) = \rho(n)z_I + \left(1 - \rho(n)\right)z_N$. Thus, although $\rho(n)$ is not explicitly provided to the GIN, the graph-level embedding depends on $\rho(n)$ through the structure of $G_n$ and the mean-pooling operation.

**Proof of Theorem 2.** The proof consists of two parts. First, using Lemma 4, we derive the general form of the SiameseGIN decision score in terms of $\rho(n)$ and $\rho(n+2)$. Second, we impose $\alpha = \beta$, reducing the score to an affine function of $D(n)$, and use Lemma 2 to establish perfect separation.

**Part 1.** By Lemma 4, applying the shared GIN to $G_n$ and $G_{n+2}$ gives

$$\begin{aligned} \mathrm{z}(\mathrm{G_n}) &= \rho(n)z_I + \left(1 - \rho(n)\right)z_N \\ \mathrm{z}(\mathrm{G_{n+2}}) &= \rho(n+2)z_I + \left(1 - \rho(n+2)\right)z_N \end{aligned}$$

The two graph-level embeddings are then concatenated to form the representation of the pair:

$$z_n = [z(G_n) \,||\, z(G_{n+2})]$$

Here, || denotes concatenation. Since the embedding vectors $z(G_n)$ and $z(G_{n+2})$ are in $R^h$ it follows that their concatenated vector $z_n$ belongs to $R^{2h}$. The final linear layer then computes the logits

$$\begin{aligned} (s_0, s_1) &= W \cdot z_n + b \\ &= W \cdot \left[\, \left(\rho(n)z_I + \left(1 - \rho(n)\right)z_N\right) \,||\, \left(\rho(n+2)z_I + \left(1 - \rho(n+2)\right)z_N\right) \,\right] + b \end{aligned}$$

where $W \in R^{2\times 2h}$ and $b \in R^2$. Since $S(n) = s_1(n) - s_0(n)$, and the final classifier is linear, collecting the terms involving $\rho(n)$, $\rho(n+2)$, and the constant terms gives

$$S_{\mathrm{GIN}}(n) = \alpha\rho(n) + \beta\rho(n+2) + \gamma$$

where $\alpha, \beta, \gamma$ are constants determined by the trained model parameters.

**Part 2.** Under the assumption $\alpha = \beta < 0$

$$\begin{aligned} S_{\mathrm{GIN}}(n) &= \alpha\rho(n) + \beta\rho(n+2) + \gamma \\ &= \alpha[\rho(n) + \rho(n+2)] + \gamma \\ &= \alpha D(n) + \gamma \end{aligned}$$

By Lemma 2 and the assumed condition on $[N_1, N_2]$,

$$\max_{\substack{n\in[N_1,N_2]\\ Positive}} D(n) < \min_{\substack{n\in[N_1,N_2]\\ Negative}} D(n)$$

Since $\alpha < 0$, multiplication by $\alpha$ reverses the inequality, and adding $\gamma$ preserves it. Hence

$$\min_{\substack{n\in[N_1,N_2]\\ Positive}} S_{\mathrm{GIN}}(n) > \max_{\substack{n\in[N_1,N_2]\\ Negative}} S_{\mathrm{GIN}}(n)$$

Therefore, every positive pair receives a strictly larger score than every negative pair, and consequently AUC = 1.

The structural representation derived in Lemma 4 is not unique to GIN. Since all vertices have identical initial features and the modular-inverse graphs consist of isolated vertices and vertices with one neighbor, the same argument extends to message-passing architectures with sum aggregation, shared node-update functions, and global mean pooling, provided that self-loops are handled consistently. We focus on GIN because it is the architecture used in our experiments, without claiming that the theoretical result is exclusive to GIN.

## 2.7 Logistic Regression - Mathematical Justification

We now investigate the theoretical capability of logistic regression using the two structural features $x_1 = \rho(n)$ and $x_2 = \rho(n+2)$. We first derive the general form of the logistic-regression decision score and show that, for an appropriate choice of weights, it reduces to an affine function of $D(n)$. Using Lemma 2, we then establish perfect separation between positive and negative instances.

**Theorem 3**. The logistic-regression decision score has the general form

$$S_{\mathrm{LR}}(n) = w_1\rho(n) + w_2\rho(n+2) + b$$

For $w_1 = w_2 = -1$, this reduces to

$$S_{\mathrm{LR}}(n) = -D(n) + b$$

Furthermore, if $n \in [N_1, N_2]$, where $N_1 > 16$ and $N_2 < \mathrm{N}_1^2/16 - 2$, then there exists a choice of $b$ for which the score perfectly separates positive and negative instances, and consequently AUC = 1.

**Experimental setting.** In our experiments, the validation interval was chosen as $N_1 = 4001$ and $N_2 = 7000$. Since $N_1 > 16$ and $7000 < 4001^2/16 - 2$, the validation interval satisfies the sufficient conditions of Lemma 2 and, consequently, those of Theorem 3. Moreover, the validation and test ranges satisfy the conditions of Lemma 3, which guarantees the existence of a common threshold $T_D$ for $D(n)$. Since $S_{\mathrm{LR}}(n) = -D(n) + b$ this induces a corresponding common threshold for the logistic-regression decision score. This is consistent with the empirical observation that the validation-selected threshold also achieved perfect separation on the test set.

The choice $w_1 = w_2 = -1$ is motivated by Lemma 2 and demonstrates the existence of a separating linear classifier. The trained logistic-regression model yielded effective weights $w_1 = -1.5002$, $w_2 = -1.8230$, and bias $b = 0.6342$. Both learned weights are negative, consistent with the theoretical motivation, although they are not equal. The theorem establishes the existence of a separating classifier rather than requiring the learned weights to equal the theoretical choice.

**Proof of Theorem 3.** The proof consists of two parts. First, we derive the general form of the logistic-regression decision score. Second, we choose $w_1 = w_2 = -1$, reducing the score to an affine function of $D(n)$, and use Lemma 2 to establish perfect separation.

**Part 1.** The logistic-regression classifier operates on the two-dimensional feature $(x_1, x_2)$ where $x_1 = \rho(n)$ and $x_2 = \rho(n+2)$, therefore its decision score has the form

$$S_{LR}(n) = w_1\rho(n) + w_2\rho(n+2) + b$$

where $w_1, w_2$ are the feature weights and $b$ is the bias.

**Part 2.** Choosing $w_1 = w_2 = -1$ we get the reduced form

$$\begin{aligned} S_{LR}(n) &= w_1\rho(n) + w_2\rho(n+2) + b \\ &= -[\rho(n) + \rho(n+2)] + b \\ &= -D(n) + b \end{aligned}$$

Since by lemma 2

$$\max_{\substack{n\in[N_1,N_2] \\ Positive}} D(n) < \min_{\substack{n\in[N_1,N_2] \\ Negative}} D(n)$$

We get

$$\max_{\substack{n\in[N_1,N_2] \\ Negative}} S_{LR}(n) < \min_{\substack{n\in[N_1,N_2] \\ Positive}} S_{LR}(n)$$

Thus, every positive instance receives a higher decision score than every negative instance. Hence, the two classes are perfectly separated and, consequently, AUC = 1. This proves the theorem.

**Experimental setting.** The choice $w_1 = w_2 = -1$ is motivated by Lemma 2, which establishes a strict separation between positive and negative instances in terms of (D(n)). This choice demonstrates the existence of a separating linear classifier. In our experiments, the trained logistic-regression model yielded effective weights $w_1 = -1.5002$ and $w_2 = -1.8230$, both negative as theoretically motivated, although not equal.

### 2.8 AUC and Perfect Threshold Separation

To relate the classification results of the SiameseGIN and logistic-regression models to their achievable balanced accuracy, we consider the relationship between perfect score ranking and threshold-based classification. The following proposition shows that an AUC of 1 guarantees the existence of a classification threshold that perfectly separates the two classes.

Let $D = \{(x_\mathrm{i}, y_\mathrm{i})\}_{i=1}^{N}$ be a finite binary classification dataset, where $y_\mathrm{i} \in \{0,1\}$, with $y_i = 1$ representing the positive class and $y_i = 0$ the negative class. Let $M$ be a binary classification model, let $s(x)$ denote the score assigned by the model to the positive class for an observation $x$.

**Proposition 1** (AUC and Perfect Threshold Separation).

If M achieves AUC = 1 on D, then there exists a threshold T that perfectly separates the two classes. Consequently, all observations are correctly classified and, in particular, BA = 1.

**Proof.** Since AUC = 1, every positive observation receives a strictly higher score than every negative observation. Therefore,

$$\max_{y=0} s(x) < \min_{y=1} s(x)$$

Hence, there exists a threshold $T$ such that

$$\max_{y=0} s(x) < T < \min_{y=1} s(x)$$

For this threshold, every observation with $y = 1$ satisfies s$(x) > T$, while every observation with $y = 0$ satisfies $s(x) \leq T$. Thus, all observations are classified correctly. Consequently, all standard classification metrics based on the resulting predictions attain their maximum value; in particular, BA = 1.

## 3 Methodology

### 3.1 Dataset

The original dataset consists of integer pairs $(n, n+2)$ constructed from integers ranging from 2 to 9999. Each pair is labeled as twin-prime if both $n$ and $n+2$ are prime, and as non-twin otherwise. The data were divided into three consecutive, non-overlapping training, validation, and test ranges, as summarized in Table 1.

**Table 1.** Distribution of twin-prime and non-twin pairs across the training, validation, and test sets.

| Set | Range | Total | Twin-prime | Non-twin | Twin % | Non-twin % |
|---|---|---|---|---|---|---|
| Train | 2-4000 | 3999 | 103 | 3896 | 2.58% | 97.42% |
| Validation | 4001-7000 | 3000 | 59 | 2941 | 1.97% | 98.03% |
| Test | 7001-9999 | 2999 | 43 | 2956 | 1.43% | 98.57% |
| **Total** | 2-9999 | 9998 | 205 | 9793 | 2.05% | 97.95% |

To further evaluate generalization, additional test sets were constructed over the numerical intervals 20,000–25,000, 40,000–45,000, 60,000–65,000, 80,000–85,000, 100,000–105,000, and 999,958–1,000,850. These sets were used exclusively for testing, without retraining or threshold adjustment. Their results are reported in Table 3.

The range-based partition was chosen to evaluate whether models trained on pairs constructed from smaller integers can generalize to pairs constructed from larger, previously unseen integers. Accordingly, all validation and test pairs lie outside the numerical range used for training. The training range $[2,4000]$ does not satisfy the conditions of Lemma 2, so perfect separation by $D(n)$ is not guaranteed over this range, which may partly explain the imperfect cross-validation results. Although the training, validation, and test sets contain distinct integer pairs, individual modular-inverse graphs may appear in adjacent splits at the range boundaries, since each sample consists of two graphs, $G_n$ and $G_{n+2}$. Consequently, although the integer pairs are disjoint across splits, the evaluation is not strictly graph-disjoint, which represents a limitation of the experimental design.

### 3.2 Graph encoding

Following the modular-inverse graph construction introduced in Section 2.1, each integer $n$ is represented by its modular-inverse graph $G_n$, whose vertex set is $V_n = \{0, 1, \ldots, n-1\}$. Two vertices $a$ and $b$ are connected whenever $ab \equiv 1(mod\, n)$. Self-loops are retained for vertices satisfying $a^2 \equiv 1(mod\, n)$. The graphs are constructed using NetworkX and converted into PyTorch Geometric data objects. Each vertex is assigned a constant one-dimensional feature $x_v = 1$, so no explicit primality information is provided through the node features. For twin-prime classification, each integer pair $(n, n+2)$ is represented by the corresponding pair of graphs $(G_n, G_{n+2})$. The pair is assigned label 1 if both $n$ and $n+2$ are prime, and label 0 otherwise. These graph pairs constitute the observations supplied to the Siamese GNN.

### 3.3 Model architectures

**SiameseGIN.** A Siamese Graph Isomorphism Network (GIN) was used to classify each graph pair $(G_n, G_{n+2})$. The two graphs were processed separately by the same GIN encoder with shared parameters. The encoder consisted of two GIN convolutional layers, each equipped with a two-layer multilayer perceptron (MLP) and ReLU activation. Following the second GIN layer, mean pooling over all nodes was applied to obtain a graph-level embedding for each graph. The resulting embeddings of $G_n$ and $G_{n+2}$ were concatenated to form a joint representation of the graph pair. The final linear classifier applied the same weight matrix to both halves of the concatenated representation, thereby enforcing equal contributions from the two graph embeddings, and produced two output logits corresponding to the non-twin and twin-prime classes. Three architectures were evaluated, with dimensions (1-8-8), (1-16-16), and (1-32-32), where the values denote the input dimension and the output dimensions of the first and second GIN layers, respectively. All architectures were trained for 10 epochs using a learning rate of 0.001 and random seed 0. Each GIN layer uses a two-layer MLP with an internal ReLU activation, followed by an additional ReLU activation after the convolution. No normalization or dropout is applied, and the GIN parameters $\varepsilon_1$ and $\varepsilon_2$ are fixed at zero. Training uses a batch size of one graph pair.

**Logistic regression**. Logistic regression was implemented using the two structural features $\rho(n)$ and $\rho(n+2)$. For each pair $(n, n+2)$, the input was the two-dimensional feature vector $(\rho(n), \rho(n+2))$, where each feature represents the proportion of isolated vertices in the corresponding modular-inverse graph. The classifier consisted of a single linear layer mapping the two input features to two output logits corresponding to the non-twin and twin-prime classes. No additional feature transformation or scaling was applied.

### 3.4 Training and model selection

**SiameseGIN.** The SiameseGIN models were trained using the Adam optimizer with a learning rate of 0.001 for 10 epochs. To account for the class imbalance between twin-prime and non-twin pairs, a weighted cross-entropy loss was used, with class weights $(1, n_0/n_1)$, where $n_0$ and $n_1$ denote the numbers of non-twin and twin-prime pairs in the training set, respectively. A random seed of 0 was used for model initialization.

**Logistic regression.** The logistic-regression model was trained using the Adam optimizer with a learning rate of 0.005 for 100 epochs and a random seed of 0. To account for class imbalance, weighted cross-entropy loss was used with the same class-weighting scheme $(1, n_0/n_1)$. The complete training feature matrix was used in each optimization step.

Following training, for both models, the output logits were converted to class probabilities using the softmax function, and the predicted probability of the twin-prime class was used as the classification score. For each model, all distinct ROC cutoffs were considered, and the threshold maximizing balanced accuracy on the validation set was selected. For the SiameseGIN, the architecture achieving the highest validation balanced accuracy, together with its corresponding validation-selected threshold, was retained for subsequent evaluation on the independent test set. The logistic-regression threshold was likewise selected exclusively on the validation set and subsequently applied unchanged to the test set.

### 3.5 Evaluation

Model performance was evaluated using accuracy, precision, recall, F1-score, balanced accuracy (BA), and area under the receiver operating characteristic curve (AUC). Balanced accuracy was used as the primary threshold-dependent performance measure because of the strong class imbalance between twin-prime and non-twin pairs. AUC was additionally used to assess the ability of the model to rank positive pairs above negative pairs independently of a particular classification threshold. After model and threshold selection on the validation set, the selected model was evaluated on the independent test set using the same validation-selected threshold, without further threshold optimization on the test data.

## 4 Results

The validation results provide the basis for model selection. All three SiameseGIN architectures achieved an AUC of 1.000 on the validation set. An AUC of 1 indicates perfect ranking of the two classes, such that every positive pair receives a higher class-1 score than every negative pair. Consequently, there exists a classification threshold that perfectly separates the two classes. For each architecture, the threshold was selected on the validation set to maximize balanced accuracy (BA), resulting in BA = 1.000. The validation results for all three SiameseGIN architectures are summarized in Table 2.

**Table 2.** Validation performance of the SiameseGIN architectures, each trained for 10 epochs using seed 0. Classification thresholds were selected on the validation set to maximize balanced accuracy.

| Arch | TP | FN | TN | FP | Acc. | Prec. | Recall | F1 | BA | AUC |
|---|---|---|---|---|---|---|---|---|---|---|
| (1-8-8) | 59 | 0 | 2941 | 0 | 1.000 | 1.000 | 1.000 | 1.000 | 1.000 | 1.000 |
| (1-16-16) | 59 | 0 | 2941 | 0 | 1.000 | 1.000 | 1.000 | 1.000 | 1.000 | 1.000 |
| (1-32-32) | 59 | 0 | 2941 | 0 | 1.000 | 1.000 | 1.000 | 1.000 | 1.000 | 1.000 |

Since all three architectures achieved perfect validation performance, the least complex architecture, (1-8-8), was selected for the subsequent test evaluation.

**Test Across Numerical Ranges**. To assess generalization beyond the original numerical range $N < 10^4$, the selected SiameseGIN (1–8–8) model was evaluated on the original test range 7,001–9,999 and six extended numerical intervals, including integers near $10^6$. The same validation-selected threshold of 0.958753 was used throughout, without retraining. The results are presented in Table 3.

**Table 3.** SiameseGIN performance across original and extended numerical ranges.

| Start | End | Twin (%) | TP | FN | TN | FP | Acc. | Prec. | Recall | F1 | BA | AUC |
|---|---|---|---|---|---|---|---|---|---|---|---|---|
| 7,001 | 9,999 | 1.4338 | 43 | 0 | 2956 | 0 | 1.00 | 1.00 | 1.00 | 1.00 | 1.00 | 1.00 |
| 20000 | 25000 | 1.3197 | 66 | 0 | 4935 | 0 | 1.00 | 1.00 | 1.00 | 1.00 | 1.00 | 1.00 |
| 40000 | 45000 | 1.0798 | 54 | 0 | 4947 | 0 | 1.00 | 1.00 | 1.00 | 1.00 | 1.00 | 1.00 |
| 60000 | 65000 | 0.8598 | 43 | 0 | 4958 | 0 | 1.00 | 1.00 | 1.00 | 1.00 | 1.00 | 1.00 |
| 80000 | 85000 | 1.1598 | 58 | 0 | 4943 | 0 | 1.00 | 1.00 | 1.00 | 1.00 | 1.00 | 1.00 |
| 100000 | 105000 | 0.9798 | 49 | 0 | 4952 | 0 | 1.00 | 1.00 | 1.00 | 1.00 | 1.00 | 1.00 |
| 999958 | 1000850 | 1.1198 | 10 | 0 | 883 | 0 | 1.00 | 1.00 | 1.00 | 1.00 | 1.00 | 1.00 |

The model achieved perfect classification across all tested intervals (BA = 1.000, AUC = 1.000), including integers near $10^6$. These results demonstrate strong empirical generalization well beyond the original training range, without retraining or threshold adjustment, and provide additional support the robustness of the proposed approach.

**Alternative Methods Comparison.** To compare SiameseGIN with alternative classification methods, we evaluated SiameseGCN, logistic regression, and the exact arithmetic baseline on the original test set (7,001–9,999). SiameseGIN and SiameseGCN were evaluated under identical experimental settings, using architecture configurations (1, 8, 8), (1, 16, 16), and (1, 32, 32), where each tuple specifies the input and two hidden dimensions. Both models were trained for 10 epochs with a learning rate of 0.001 and random seed 0. Model selection and classification-threshold optimization were performed exclusively on the validation set. As established in Lemma 2, the features $\rho(n)$ and $\rho(n+2)$ admit perfect linear separation between twin-prime and negative pairs under the stated bounded-range conditions. The exact arithmetic baseline introduced in Section 2.3 requires neither training nor threshold optimization. The results are summarized in Table 4.

**Table 4.** Test performance comparison of SiameseGIN, SiameseGCN, logistic regression, and the exact arithmetic baseline on the original test set (7,001–9,999).

| Model | Threshold | TP | FN | TN | FP | Acc | BA | AUC |
|---|---|---|---|---|---|---|---|---|
| SiameseGIN | 0.958753 | 43 | 0 | 2956 | 0 | 1.000 | 1.000 | 1.000 |
| Logistic Regression | 0.653257 | 43 | 0 | 2956 | 0 | 1.000 | 1.000 | 1.000 |
| Exact arithmetic baseline | N/A | 43 | 0 | 2956 | 0 | 1.000 | 1.000 | 1.000 |
| SiameseGCN | >1 | 0 | 43 | 2956 | 0 | 0.986 | 0.500 | 0.500 |

SiameseGIN, logistic regression, and the exact arithmetic baseline achieved perfect classification. In contrast, SiameseGCN achieved a balanced accuracy of 0.500 and an AUC of 0.500 under the evaluated configurations. Its validation-selected threshold exceeded 1, resulting in exclusively negative predictions. Thus, SiameseGIN outperformed SiameseGCN in the present experiment.

**Cross Validation.** To evaluate generalization across numerical ranges, five-fold range-based cross-validation was performed on the combined training and validation datasets, excluding the independent test set. The samples were sorted by increasing $n$ and divided into five consecutive, non-overlapping folds. In each iteration, one fold was reserved exclusively for testing, while the remaining four folds were divided into training (80%) and validation (20%) subsets using stratified random sampling. A new SiameseGIN model was trained for each iteration, and the classification threshold was selected exclusively from the corresponding validation subset. The held-out fold was used exclusively for testing and was not involved in model training or threshold calculation. The SiameseGIN (1-8-8) architecture was fixed across all five folds, with 10 training epochs, a learning rate of 0.001, and random seed 0. The results are summarized in Table 5.

**Table 5.** Five-fold range-based cross-validation results for SiameseGIN, with thresholds selected on independent validation sets.

| Fold | Threshold | BA | AUC |
|---|---|---|---|
| 1 | 0.9709 | 0.8804 | 0.9864 |
| 2 | 0.9263 | 0.9993 | 1.0000 |
| 3 | 0.8775 | 0.9898 | 1.0000 |
| 4 | 0.9446 | 0.9989 | 1.0000 |
| 5 | 0.9737 | 1.0000 | 1.0000 |
| **Mean ± Std** | - | 0.9737 ± 0.0523 | 0.9973 ± 0.0061 |

The five-fold cross-validation yielded a mean balanced accuracy of 0.9737 ± 0.0523 and a mean AUC of 0.9973 ± 0.0061. Fold 1 exhibited the lowest balanced accuracy (0.8804), despite achieving a high AUC (0.9864). This may be attributed to the difference in numerical ranges between training and testing, as Fold 1 contains the smallest numbers, while the model was trained on larger numbers. Consequently, the validation-selected threshold may not transfer perfectly to the smaller numerical range, resulting in 11 false negatives and no false positives. The remaining folds achieved balanced accuracies between 0.9898 and 1.0000. Overall, the results indicate strong discrimination across numerical ranges, with some sensitivity to threshold selection.

**Multi-Seed Analysis.** To evaluate the robustness of SiameseGIN to random initialization, we repeated the experiment using five different random seeds (1–5). The seeds generated different initial model weights, while the data partition and all other experimental settings remained fixed. In each run, the classification threshold was selected exclusively on the validation set and applied unchanged to the independent test set. The results are presented in Table 6.

**Table 6.** SiameseGIN performance on the independent test set across five random seeds.

| Seed | TP | FN | TN | FP | ACC | Precision | Recall | F1 | BA | AUC |
|---|---|---|---|---|---|---|---|---|---|---|
| 1 | 43 | 0 | 2956 | 0 | 1.00 | 1.00 | 1.00 | 1.00 | 1.00 | 1.00 |
| 2 | 43 | 0 | 2956 | 0 | 1.00 | 1.00 | 1.00 | 1.00 | 1.00 | 1.00 |
| 3 | 43 | 0 | 2956 | 0 | 1.00 | 1.00 | 1.00 | 1.00 | 1.00 | 1.00 |
| 4 | 43 | 0 | 2956 | 0 | 1.00 | 1.00 | 1.00 | 1.00 | 1.00 | 1.00 |
| 5 | 43 | 0 | 2956 | 0 | 1.00 | 1.00 | 1.00 | 1.00 | 1.00 | 1.00 |

All five runs achieved perfect classification, with a mean balanced accuracy of 1.0000 ± 0.0000 and a mean AUC of 1.0000 ± 0.0000. This demonstrates consistent performance across the five different initializations, indicating robustness with respect to the initial model weights.

## 5 Discussion

The experimental results are consistent with the theoretical analysis presented in this work. Both the SiameseGIN and logistic regression achieved perfect separation on the independent test set, showing that the modular-inverse graphs contain strong structural information for distinguishing twin-prime from non-twin pairs. Furthermore, the model achieved perfect classification across all extended numerical ranges, including integers near $10^6$, without retraining or threshold adjustment (Table 3).

The mathematical characterization of the SiameseGIN provides an explanation for this behavior. As established in Lemma 4 and Theorem 2, the graph-level representations depend on the isolated-vertex proportions $\rho(n)$ and $\rho(n+2)$, and the resulting decision score can be expressed as an affine function of their sum. The logistic-regression results further confirm that this structural information is sufficient for perfect classification under the established separation conditions.

An important distinction between the two approaches is how the structural information is obtained. Logistic regression uses the isolated-vertex proportions as explicitly computed input features, whereas the SiameseGIN extracts this information directly from the modular-inverse graphs through message passing and mean pooling. Although both approaches achieved perfect classification on the evaluated test sets, the SiameseGIN demonstrates how a graph-learning architecture can represent the relevant structural information without requiring these proportions as input features. Its contribution is therefore representational rather than an improvement in classification accuracy over logistic regression.

The exact arithmetic baseline further demonstrates that the isolated-vertex proportions encode sufficient information to determine twin primality directly. Since this deterministic rule guarantees perfect classification, the contribution of SiameseGIN is not to outperform an exact number-theoretic criterion, but to characterize how a graph neural network represents and processes the relevant structural information.

An important result is that, in the original experimental split, the classification thresholds were selected on the validation set and then applied unchanged to the independent test set, yielding perfect classification. This result is consistent with Lemma 3, which establishes conditions under which the validation and test ranges admit a common separating threshold. However, the additional range-based cross-validation reveals some variation in threshold transfer across numerical ranges. In particular, Fold 1 achieved a high AUC (0.9864) but a lower balanced accuracy (0.8804). This may be attributed to the difference in numerical ranges between training and testing, as Fold 1 contains the smallest numbers, while the model was trained on larger numbers. Consequently, the validation-selected threshold may not transfer perfectly to the held-out range. This distinction highlights the difference between ranking performance and threshold-dependent classification.

These results are limited to the finite numerical ranges considered in the experiments and do not imply a general result about twin primes for arbitrary integers. The theoretical analysis establishes sufficient conditions for perfect classification, while the experiments demonstrate that the SiameseGIN achieves the predicted behavior on the considered numerical ranges. The principal contribution is therefore the mathematical characterization of how the SiameseGIN architecture processes this structure, with the experiments demonstrating agreement with the theoretical analysis.

## 6 Limitations

The experiments were conducted over finite and predefined numerical ranges, and therefore the observed perfect separation should not be interpreted as a general result for arbitrarily large integers. In addition, the study considers a specific modular-inverse graph representation and a limited set of GIN architectures. The theoretical results provide sufficient separation conditions under the stated assumptions and ranges, rather than a general statement about all twin-prime pairs.

## 7 Future work

Future work may extend the experiments to numerical ranges beyond those evaluated in this study to examine whether the observed separation and threshold transfer continue to hold. It would also be useful to investigate additional graph representations and GNN architectures.

Finally, further theoretical analysis could study whether stronger separation results can be established beyond the bounded numerical ranges considered in this work.

## 8 Conclusions

This study investigated twin-prime classification using pairs of modular-inverse graphs. The principal contribution is a mathematical characterization of how the SiameseGIN architecture processes these graphs, showing that its graph-level representations depend on the isolated-vertex proportions $\rho(n)$ and $\rho(n + 2)$, and that its decision score can be expressed as an affine function of their sum. The theoretical analysis established sufficient conditions for separating twin-prime and non-twin pairs, while the experiments showed that both the SiameseGIN and logistic regression achieved perfect classification on the independent test set using validation-selected thresholds. The agreement between the theoretical and experimental results demonstrates how the structure of modular-inverse graphs determines the classification behavior of the considered models within the investigated numerical ranges.

## Declarations

### Data Availability Statement

The data used in this study were generated computationally from the integers considered in the experiments using the modular-inverse graph construction described in the manuscript. The data and code used to generate the graphs and reproduce the experiments are available from the corresponding author upon reasonable request.

### Competing interests

The author declares no competing interests.

### Funding

The author received no funding for this work.

### Authors' contributions

The sole author was responsible for the conception, methodology, theoretical analysis, experiments, analysis of results, and preparation of the manuscript.

### Use of Generative-AI Tools Declaration

ChatGPT (OpenAI) was used to assist with language editing. The author reviewed and verified the final content and takes full responsibility for the manuscript.

### Acknowledgements

Not applicable.